\documentclass{informs3}%
\DoubleSpacedXI%
\usepackage{tikz}%
\usepackage{siunitx}%
\usepackage{stfloats}%
\usepackage{enumitem}%
\setlist[enumerate]{itemsep=.2mm,topsep=1.5mm}%
\setlist[itemize]{itemsep=.2mm,topsep=1.5mm}%
\newcommand{\ignore}[1]{}%
\usepackage{xspace}%
\usepackage{amsmath}%
\newcommand{\MVC}{\textsc{MVC}\xspace}%
\newcommand{\TSP}{\textsc{TSP}\xspace}%
\newcommand{\fastvc}{\textsc{FastVC}\xspace}%
\renewcommand{\bar}{\textsc{Bet-and-Run}\xspace}%
\newcommand{\luby}{\textsc{Luby}\xspace}%
\newcommand{\even}{\textsc{Even}\xspace}%
\usepackage{times}%
\usepackage{xcolor}%
\usepackage{soul}%
\usepackage[utf8]{inputenc}%
\usepackage[small]{caption}%
\usepackage{latexsym}%
\usepackage{graphicx}%
\usepackage{bm}%
\usepackage{hyperref}%
\makeatletter%
\gdef\budget#1{\ensuremath{\if\relax\detokenize{#1}\relax\expandafter\@firstoftwo\else\expandafter\@secondoftwo\fi{t}{t_{#1}}}}
\gdef\totalBudget{\budget{}}
\gdef\initialRuns{\ensuremath{k}}
\gdef\selectedRuns{\ensuremath{m}}
\gdef\initialBudget{\budget{1}}%
\gdef\initialRunBudget#1{\ensuremath{\budget{1,#1}}}%
\gdef\decisionMaker{\ensuremath{D}}%
\gdef\decisionMakerTime{\budget{2}}%
\gdef\selectedAdditionalBudget{\budget{3}}%
\gdef\inQuotes#1{``#1''}%
\protected\gdef\dmStyle#1{\textsf{#1}}%
\gdef\currentBest{\dmStyle{currentBest}}%
\gdef\currentWorst{\dmStyle{currentWorst}}%
\gdef\random{\dmStyle{random}}%
\gdef\mostImprovements{\dmStyle{mostImprovements}}%
\gdef\logTimeSum{\dmStyle{logTimeSum}}%
\gdef\diminishingReturns{\dmStyle{diminishing}}%
\gdef\perceptron#1{\dmStyle{PER(\ensuremath{#1})}}%
\gdef\npPrefix{\ensuremath{\mathcal{NP}}}%
\gdef\npHard{\textnormal{\npPrefix-hard}}%
\gdef\instanceStyle#1{\texttt{#1}}%
\gdef\citeAsName#1{\citeauthor{#1} [\citeyear{#1}]}%
\makeatother%
\usepackage{endnotes}%
\let\footnote=\endnote%
\let\enotesize=\normalsize%
\def\notesname{Endnotes}%
\def\makeenmark{\ensuremath{^{\theenmark}}}%
\def\enoteformat{\rightskip0pt\leftskip0pt\parindent=1.75em\leavevmode\llap{\theenmark.\enskip}}%
\usepackage{natbib}
\bibpunct[, ]{(}{)}{,}{a}{}{,}%
\def\bibfont{\small}%
\TheoremsNumberedThrough
\ECRepeatTheorems%
\EquationsNumberedThrough
\begin{document}%
%
%
\RUNAUTHOR{Weise, Wu, and Wagner}%
%
\RUNTITLE{An Improved Generic \bar Strategy for Speeding Up Stochastic Local Search}%
\TITLE{An Improved Generic \bar Strategy for Speeding Up Stochastic Local Search}%
%
\ARTICLEAUTHORS{%
\AUTHOR{Thomas Weise}%
\AFF{Institute of Applied Optimization, Faculty of Computer Science and Technology, Hefei University, \EMAIL{tweise@hfuu.edu.cn}}%
\AUTHOR{Zijun Wu}%
\AFF{Institute of Applied Optimization, Faculty of Computer Science and Technology, Hefei University, \EMAIL{zijunwu1984a@163.com}}%
\AUTHOR{Markus Wagner}%
\AFF{Optimisation and Logistics, The University of Adelaide, Adelaide, Australia, \EMAIL{markus.wagner@adelaide.edu.au}}%
}
\ABSTRACT{%
A commonly used strategy for improving optimization algorithms is to restart the algorithm when it is believed to be trapped in an inferior part of the search space. 
Building on the recent success of \bar approaches for restarted local search solvers, we introduce an improved generic \bar strategy. 
The goal is to obtain the best possible results within a given time budget \totalBudget\ using a given black-box optimization algorithm.
If no prior knowledge about problem features and algorithm behavior is available, the question about how to use the time budget most efficiently arises. We propose to first start $\initialRuns\geq 1$ independent runs of the algorithm during an initialization budget $\initialBudget<\totalBudget$, pausing these runs, then apply a decision maker \decisionMaker\ to choose $1\leq\selectedRuns<\initialRuns$ runs from them (consuming $\decisionMakerTime\geq 0$ time units in doing so), and then continuing these runs for the remaining $\selectedAdditionalBudget=\totalBudget-\initialBudget-\decisionMakerTime$ time units.
In previous \bar strategies, the decision maker $\decisionMaker=\currentBest$ would simply select the run with the best-so-far results at negligible time.
We propose using more advanced methods to discriminate between ``good'' and ``bad'' sample runs, with the goal of increasing the correlation of the chosen run with the a-posteriori best one. We test several different approaches, including neural networks trained or polynomials fitted on the current trace of the algorithm to predict which run may yield the best results if granted the remaining budget. 
We show with extensive experiments that this approach can yield better results than the previous methods, but also find that the \currentBest\ method is a very reliable and robust baseline approach.}%
%
%
\KEYWORDS{Restart strategies, algorithm performance, heuristic search, erraticism}%
\HISTORY{This paper was first submitted in June 2018.}%
\maketitle%
%
%
%
%
\section{Introduction}%
Optimization algorithms are widely used in a variety of domains, such as production scheduling and planning or vehicle routing.
In many such practical applications, the total time budget \totalBudget\ available for optimization is limited to at most a few minutes.
The goal is to find a solution which is as-good-as-possible within this budget.
One method to do so is to develop better optimization algorithms.
Another method is to make the best use of an existing solver.

There are two straightforward methods when approaching an optimization problem with one algorithm and a total time budget \totalBudget: One can either assign the whole budget \totalBudget\ to a single run of the algorithm or execute \initialRuns\ independent restarts of the algorithm~\citep{M2003MSM,LMS2010ILSFAA} and therefore divide the budget \totalBudget\ into \initialRuns\ equally-sized chunks.
It can be expected that the former strategy is the best for small budgets while the latter one is the better choice for large budgets.
Budgets \totalBudget\ of a few minutes, however, fall in neither category for many problem types, which, of course, depends on the instance and the solver.

Here, \bar strategies~\citep{FM2014EEIS} pose a compromise by using an initialization time budget $0\leq\initialBudget\leq\totalBudget$ which is divided evenly amongst \initialRuns\ independent runs.
The run with the best-so-far solution is then continued for the remaining $\totalBudget-\initialBudget$ time units. \citeAsName{FKW2017AGBARSFSUSLS}~showed recently that this simple approach can routinely outperform the two budgeting approaches above.
Yet, it makes only use of a single unit of information per run for the \inQuotes{decision} which of the \initialRuns\ runs to resume, namely the solution quality they reached at the end of their respective initialization budgets.
 
\begin{figure}[tb]%
\centering\includegraphics[width=0.75\columnwidth]{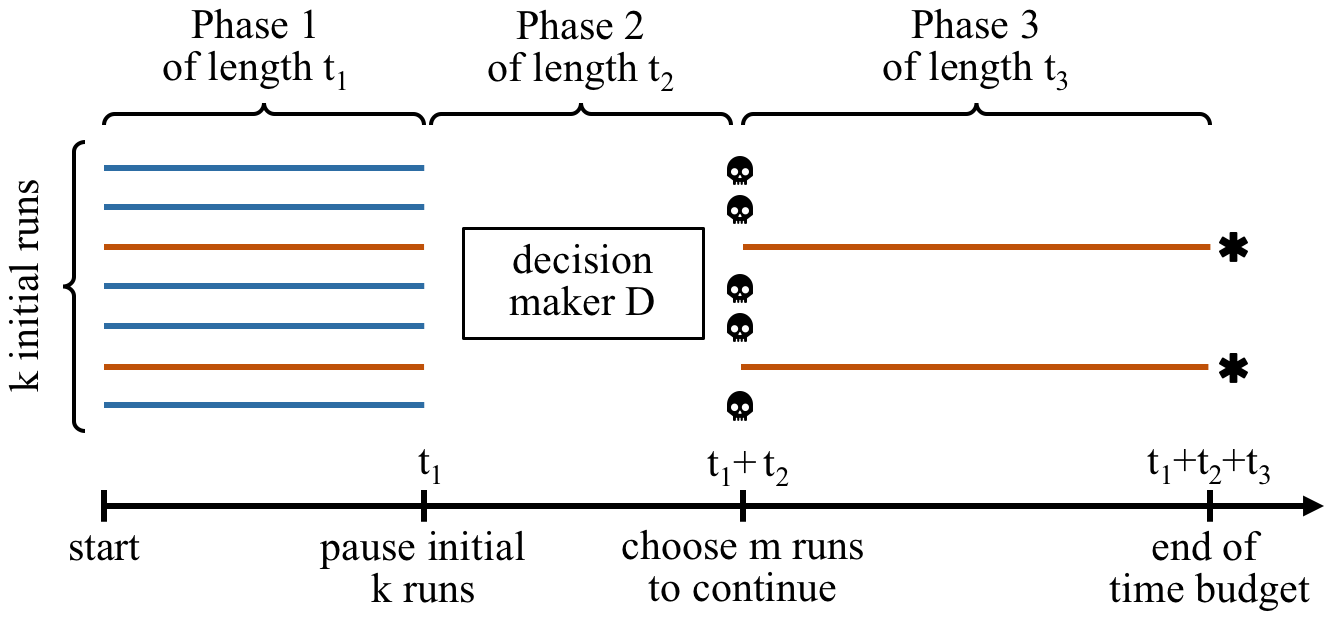}
\caption{Our generic \bar\ strategy receives a total time budget \totalBudget. It starts \initialRuns\ independent runs and pauses them after time \initialBudget. A decision maker \decisionMaker\ then takes time \decisionMakerTime\ to decide which of them to continue. All but the \selectedRuns\ chosen runs are terminated (marked with
\protect\includegraphics[height=2mm]{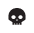}).
The \selectedRuns\ chosen runs (marked with
\protect\includegraphics[height=2.3mm]{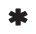})
continue for a total of \selectedAdditionalBudget.%
}%
\label{fig:illustration}%
\end{figure}

In order to investigate the question \emph{Can we do better than that?}, we generalize the \bar\ concept as illustrated in Figure~\ref{fig:illustration}:
The budget \totalBudget\ is divided into three pieces, i.e., $\totalBudget=\initialBudget+\decisionMakerTime+\selectedAdditionalBudget$ and used as follows.%
\begin{description} %
\item[Phase 1] The initialization budget \initialBudget\ is divided among a set of \initialRuns\ initial, independent runs according to a budgeting strategy. %
All of these runs are paused after \initialBudget\ has been consumed.%
\item[Phase 2] Then, a decision maker \decisionMaker\ is applied which may access the history of each run in form of $(\textit{time}, \textit{quality})$ tuples.
\decisionMaker~will choose $1\leq\selectedRuns\leq\initialRuns$ of the \initialRuns\ runs for continuation and consume an \emph{a priori} unknown time \decisionMakerTime\ while doing so.%
\item[Phase 3] The remainder \selectedAdditionalBudget\ of the total budget \totalBudget\ is then divided evenly among the \selectedRuns\ chosen runs, which thus each receive $(\totalBudget-\initialBudget-\decisionMakerTime)/\selectedRuns$ additional time units.%
\end{description}%

In the following, we show that our approach can yield an advantage above the two straightforward methods for solving optimization problems mentioned at the beginning of the introduction, and it also often outperforms the \bar strategy by~\citeAsName{FKW2017AGBARSFSUSLS} when this is possible. 

In this article, we first survey existing work in Section~\ref{sec:related} and introduce the used dataset in Section~\ref{sec:benchmarks}. Then, we motivate the use of different decision makers in Section~\ref{sec:ibar} before we present the results of our comprehensive study in Section~\ref{sec:experiments}. All datasets and the complete implementation of our algorithms have been made publicly available~\citep{Weise2018barCode}.%
%
%
\section{Related Work}\label{sec:related}%
In practice, stochastic search algorithms and randomized search heuristics are frequently restarted: If a run does not conclude within a pre-determined solution quality limit, we restart the algorithm~\citep{M2003MSM,LMS2010ILSFAA}. 
One of the advantages of this simple approach is that it helps to avoid heavy-tailed runtime distributions~\citep{GomesSCK00}.
However, due to the added complexity of designing an appropriate restart strategy for a given target algorithm, the two most common techniques used are to either restart with a certain probability at the end of each iteration, or to employ a fixed schedule of restarts.

Some theoretical results exist on how to construct optimal restart strategies. For example, \citeAsName{LSZ1993OSOLVA} showed that, for Las Vegas algorithms with known run time distribution, there is an optimal stopping time in order to minimize the expected running time. Even if the distribution is unknown, there is a universal sequence of running times given by (1,1,2,1,1,2,4,1,1,2,1,1,2,4,8,\dots), 
which is the optimal restarting strategy up to constant factors. 
While these results can be used for every problem setting, they only apply to Las Vegas algorithms. 

Fewer results are known for the optimization case. Introductions to practical approaches for such restart strategies are given by
\citeAsName{M2003MSM} and \citeAsName{LMS2010ILSFAA}. A relatively recent theoretical result is presented by \cite{Sch-Tey-Tey:c:12}. Several studies show the substantial impact of the restart policy on the efficiency of solvers for satisfiability problems~\citep{Biere2015evaluatingRestarts,Huang2007}. In this context, restarts have also been used to learn ``no-goods'' during backtracking~\cite{CireKS14}.

Quite often, classical optimization algorithms are deterministic and thus cannot be improved by restarts, as run time and outcome will not change. 
However, their characteristics can be subject to change. For example, \citeAsName{Lalla-Ruiz2016} exploited this by using different mathematical programming formulations so as to provide different starting points for the solver. 
While many other modern optimization algorithms also work mostly deterministically, they often have some randomized component, for example by choosing a random starting point. 
These initial solutions often strongly influence the quality of the outcome and the speed of reaching it. In our opinion, it follows quite naturally that algorithms should be run several times.

\citeAsName{FM2014EEIS} extended the classical restart strategies to the so-called \bar strategy:%
\begin{description}%
\item[Phase 1] perform \initialRuns\ runs of the algorithm for some (short) time limit $\initialBudget\leq\totalBudget$, assigning $\initialBudget/\initialRuns$ time units to each run.%
\item[Phase 2] use remaining time $\selectedAdditionalBudget=\totalBudget-\initialBudget$ to continue \emph{only the best run} from the first phase until timeout.%
\end{description}%

\citeAsName{FM2014EEIS} experimentally studied this for mixed-integer programming. They explicitly introduce diversity in the starting conditions of the used MIP solver (IBM ILOG CPLEX) by directly accessing internal mechanisms. For them, $\initialRuns=5$ performed best.

\citeAsName{dePerthuisdeLaillevault2015onemaxInits} have shown that a \bar strategy can also benefit asymptotically from larger \initialRuns. For the pseudo-boolean test function \textsc{OneMax} it was proven that choosing $\initialRuns>1$ decreases the $O(n \log n)$ expected run time of the \mbox{(1+1)}~evolutionary algorithm by an additive term of $\Omega(\sqrt{n})$.

\citeAsName{Lissovoi017theory} investigated \bar for a family of pseudo-Boolean functions, consisting of a plateau and a slope, as an abstraction of real fitness landscapes with promising and deceptive regions. 
The authors proved that \bar with non-trivial \initialRuns\ and \initialBudget\ are necessary to find the global optimum efficiently, and that the choice of \initialBudget\ is linked to features of the problem. They also provided a fixed budget analysis to guide selection of the \bar parameters to maximize the solution quality.

\citeAsName{FKW2017AGBARSFSUSLS} investigated a comprehensive range of \bar strategies on the traveling salesperson problem and the minimum vertex cover problem. 
Their best strategy performed 40~short runs in the initial phase with a time limit that is 1\% of the total time budget each, and then it used the remaining 60\% of the total time budget to continue the best run of the first phase. They investigated the use of the universal sequence of \citeAsName{LSZ1993OSOLVA} as well, using various choices of $t_1$, however, it turned out inferior.

Building on the success of \bar approaches for restarted local search solvers, \citeAsName{KSW2017LARRSTISS} introduced the idea of adaptive restart strategies. Inside their approach, a learned black-box decision procedure \ignore{(learned via automated algorithm configuration~\cite{AnsoteguiST09gga}) }dynamically decides whether to continue the current run, to continue a previous run, or to start a new run. While their approach performed favorably, the internal mechanisms were black-box and it remained unclear which algorithmic components and which decisions contributed to the success.

Note that the stream of \bar-related research is related to the very mature field of multi-armed bandits. To the best of our knowledge, however, there are no existing works there that make use of the core ideas of \bar to solve a single instance, i.e., to have an overall limited budget as well as the idea to exclusively stick to one arm after some first exploratory phase. For example, \citeAsName{Gagliolo2011} propose a method for allocating computation time to algorithm portfolios for solving instance sets (thus working on a much higher/coarser granularity), however, their approach does not carry over to our fine-grained scenario of using partial runs for optimizing a single instance.
%
%
\section{Benchmarks}%
\label{sec:benchmarks}%
Here we shortly introduce the two benchmark problems and the optimization algorithms used to solve them.%
\subsection{Minimum Vertex Cover Problem}%
Solving the minimum vertex cover problem (\MVC) means finding the smallest set of vertexes of a graph which contains at least one vertex from every edge.
The \MVC\ is one of the classical \npHard\ problems with many applications~\citep{GMPV2006EAOAAFTVCASCP}.
It also is closely related to the problem of finding a maximum clique~\cite{KLSS2006SPAFFP}. The state-of-the-art algorithms for solving the \MVC\ comprise \fastvc~\citep{C2015BBCAQLSFMVCIMG}, NuMVC~\citep{CSLS2013NAELSAFMVC}, TwMVC~\citep{CLS2015TWLSFMVC}, and FastWVC~\citep{LCH2017AELSAFMWVCOMG}.

\citeAsName{KSW2017LARRSTISS} applied \fastvc~\cite{C2015BBCAQLSFMVCIMG} in their experiments, one of the best algorithms for large \MVC\ instances.
\fastvc\ is based on two low-complexity heuristics. The first one constructs an initial vertex cover and the second one chooses the vertex to be removed in each exchanging step, which involves random draws from a set of candidates.
We use the data that \citeAsName{KSW2017LARRSTISS} gathered in 10,000 independent runs on all 86 instances used by~\cite{C2015BBCAQLSFMVCIMG}. 
These instances are of rather large scale and most of them are sparse, which is challenging for solvers. The number of vertices in the instances ranges from about 1000 to over 4 million and the number of edges from about 2000 to over 56 million.%
\subsection{Traveling Salesperson Problem}%
The traveling salesperson problem (\TSP)~\citep{ABCC2006TTSPACS,LLKS1985TTSPAGTOCO} is one of the most well-known combinatorial optimization tasks.
A \TSP\ instance is defined as a fully-connected graph.
Each edge in the graph has a weight, representing the distance between the two nodes it connects.
A candidate solution is a cycle that visits each node in the graph exactly once and returns back to its starting node.
The objective function, subject to minimization, is the sum of the weights of all edges in the tour, i.e., the total tour length. This optimization version of the \TSP\ is \npHard~\citep{G1979CAIAGTTTONC,GP2004TTSPAIV}. The state-of-the-art algorithms for the \TSP\ include EAX~\citep{NK2013APGAUEACFTTSP}, the Chained-Lin-Kernighan heuristic~\citep{ACR2003CLKFLTSP,C2005TTSPDW}, Partition Crossover~\citep{W2016BNMDPCADIM}, as well as hybrid metaheuristics~\citep{LWWC2015HECMFTTSP}.

We use the data from \citeAsName{KSW2017LARRSTISS}, who used the Chained-Lin-Kernighan heuristic~\citep{ACR2003CLKFLTSP,C2005TTSPDW} as \TSP\ solver.
They applied it 10,000 times to each of the 112 instances from TSPLib~\cite{R1991ATSPL} and additionally to the large instances \instanceStyle{ch71009}, \instanceStyle{mona-lisa100k}, and \instanceStyle{usa115475}.
We omit instance \instanceStyle{linhp318}, as no data was available on it.

In the next sections, for the sake of readability, we will refer to the combination of solver and problem by just using the problem domain, i.e., \TSP and \MVC.
%
%
\section{\bar\ with Better Decision Makers}%
\label{sec:ibar}%
\label{sec:sub:selection}%
\definecolor{cinitialBest}{RGB}{0,147,221}%
\definecolor{cendBest}{RGB}{151,69,120}%
\definecolor{cinitBudget}{RGB}{218,37,29}%
\definecolor{cendBudget}{RGB}{0,146,63}%
Our generalized \bar\ strategy can simulate a range of existing approaches. 
For instance, the simple multi-run strategy of restarting from scratch \initialRuns\ times is a special case by choosing $\initialBudget=\totalBudget/\initialRuns$ and $\decisionMakerTime=\selectedAdditionalBudget=0$.
The single-run strategy corresponds to a multi-run method with $\initialRuns=1$.
The strategies from \citeAsName{FM2014EEIS,FKW2017AGBARSFSUSLS} and \citeAsName{Lissovoi017theory} are special cases by choosing $\selectedRuns=1$ and having $\decisionMakerTime\approx0$.
Our experiments detailed in the next section therefore also cover these approaches. 

\begin{figure}[tb]%
\centering%
\includegraphics[width=0.4\columnwidth]{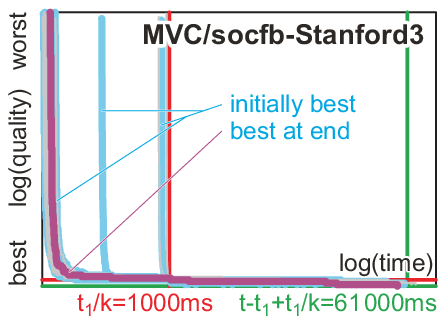}%
\hfill%
\includegraphics[width=0.4\columnwidth]{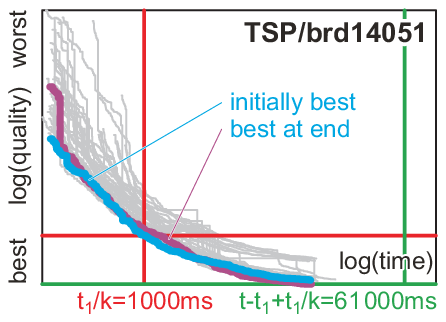}%
\caption{$\initialRuns=40$ selected runs from the datasets \instanceStyle{brd14051} (TSP) and \instanceStyle{socfb-Stanford3} (MVC) for a total budget of $\totalBudget=100'000ms$ and an initialization budget of $\initialBudget=40'000ms$, illustrating that the \textcolor{cinitialBest}{runs which are best after the initialization budget} (\textcolor{cinitBudget}{$\initialBudget=1000ms$}) are not necessarily the \textcolor{cendBest}{best ones after the full budget} (\textcolor{cendBudget}{$\initialBudget+\selectedAdditionalBudget=61000ms$}).}\label{fig:10kruns}%
\end{figure}

Note that in all related work, $\decisionMaker=\currentBest$ is applied, which picks one of the \textcolor{cinitialBest}{runs with the current best result} after initialization and resumes it where it was paused. Intuitively, this is a robust strategy for which $\decisionMakerTime\approx 0$ holds, but it does not necessarily pick the \textcolor{cendBest}{run which yields the best result} after all budget has been exhausted, as illustrated in Figure~\ref{fig:10kruns}. In such scenarios, paying some time cost \decisionMakerTime\ for a more sophisticated choice could yield a better overall result.

In our experiments, we need to use clock time as time measure and cannot apply any other measure common in optimization~\citep{WCTLTCMY2014BOAAOSFFTTSP} such as function evaluations (FEs). This is because the decision makers do not evaluate the objective function or generate candidate solutions by themselves, but only process the data already collected, i.e., the aforementioned $(\textit{time}, \textit{quality})$ tuples. Using clock time to measure the computational effort of both the optimization algorithm and the decision maker has the further advantage that we can also consider otherwise ``hidden'' costs, such as for the initialization of data structures and bookkeeping.

In order to get an initial estimate on how likely such scenarios are, we randomly draw 1'000'000 samples of \initialRuns\ runs from our each of our 86 \MVC\ and 113 \TSP\ datasets. In Table~\ref{tbl:scenarioChanges}, we count how often the runs chosen by \currentBest, which were best after time $\initialBudget/\initialRuns$, are outperformed by another run after time $\totalBudget-\initialBudget+\initialBudget/\initialRuns$. %
\begin{table}%
\centering%
\caption{Baseline performance of \currentBest. Shown is the number (and percentage) of instances from the \MVC\ and \TSP\ experiments where another decision maker could potentially outperform \currentBest\ and the estimated overall probability averaged over all datasets in an experiment for $\totalBudget=100s$ and $\initialBudget=40s$.}%
\label{tbl:scenarioChanges}%
\centering%
\begin{tabular}{lrcc}%
\hline%
experiment&\initialRuns&instances&mean probability\\%
\hline%
\MVC&4&67 (78\%)&0.11\\%
\MVC&10&52 (60\%)&0.17\\%
\MVC&40&32 (37\%)&0.23\\%
\TSP&4&30 (27\%)&0.03\\%
\TSP&10&34 (30\%)&0.06\\%
\TSP&40&44 (39\%)&0.12\\%
\ignore{
\MVC&4&67&78\%&0.11\\%
\MVC&10&52&60\%&0.17\\%
\MVC&40&32&37\%&0.23\\%
\TSP&4&30&27\%&0.03\\%
\TSP&10&34&30\%&0.06\\%
\TSP&40&44&39\%&0.12\\%
}
\hline%
\end{tabular}%
\end{table}

As can be seen, at least for $\totalBudget=100s$ and $\initialBudget=40s$, \currentBest\ cannot be beaten in the majority of samples. Still, in 78\% of the \MVC\ benchmark instances, there were at least some samples of $\initialRuns=4$ runs where \currentBest\ made the wrong choice. For $\initialRuns=40$, the chance to theoretically being able to outperform \currentBest\ on a random instance of the \MVC\ problem is 23\%. For the \TSP, these chances tend to be lower, but there is still a potential to improve the overall performance. However, these are mean probabilities, and the actual values can deviate significantly. For example, for the two instances shown in Figure~\ref{fig:10kruns}, the observed probabilities of outperforming \currentBest\ when varying $\initialRuns\in\left\{4,10,40\right\}$ range from 0.25 to 0.93 (\instanceStyle{brd14051}) and 0.01 to 0.10 (\instanceStyle{socfb-Stanford3}).

A decision maker \emph{better} than \currentBest\ would need to, e.g., outperform it in at least some of these scenarios while not performing worse in others. Although we have confirmed that there exist sufficiently many scenarios where this is potentially possible, there is another requirement which may decrease these chances: The performance data collected during the initialization budget \initialRunBudget{i} of a run $i$ must permit making a sufficiently accurate prediction regarding the future progress of that run. If this is true, then sophisticated decision makers have a chance to yield better results. Qi et~al., for instance, showed that perceptrons have good prediction accuracy in their experiments on the Maximum Satisfiability Problem~\cite{QWL2017MOARBAIA} and the TSP~\cite{QWL2018OABMASOTTSP} with simple solvers. This prediction capability should make them suitable for determining which solution quality a run would yield if continued for a certain amount of time. But other techniques, such as linear predictors, might yield improvements as well. An investigation of different \bar\ configurations and decision makers therefore is worthwhile.%
%
%
%
\section{Experimental Study}%
\label{sec:experiments}%
\subsection{Experimental Setup}%
To investigate the performance of our approach, and in particular to investigate the benefits of our more general \bar setup, we first perform a wide scan of many different setups and then investigate fewer setups in more detail. All datasets have been made publicly available~\citep{Weise2018barCode}.%
\subsubsection{Initial Large-Scale Experiment.}%
\label{sec:initialExperiment}%
In the first set of experiments, we limit ourselves to 20 random samples for each benchmark dataset and setup. We cover the total budgets $\totalBudget\in\left\{1s,4s,10s,40s,100s,400s\right\}$. For $\initialRuns\in\left\{4,10,40\right\}$ and $\selectedRuns\in\left\{1,2\right\}$, we test 25 different values of \initialBudget. These are automatically chosen according to a heuristic based on the data from \cite{KSW2017LARRSTISS} for each instance before the experiment in order to maximize the number of different, meaningful outputs, e.g., the smallest values of \initialBudget\ are chosen that there is at least one data point. We briefly investigated more choices for \selectedRuns, but found that the preliminary results did not look very promising while the overall experimental time required would have more than doubled.

For the distribution of \initialBudget\ among the \initialRuns\ initial runs, two strategies are tested. \even assigns $\initialRunBudget{i}=\initialBudget/\initialRuns$ for each $i\in 1..\initialRuns$. \luby\ instead follows the Luby sequence~\cite{LSZ1993OSOLVA} and sets $\initialRunBudget{i}=l(i)$, which equals to $2^{z-1}$ if $i=2^z - 1$ and to $i-2^{z-1}+1$ otherwise, with $2^{z-1}\leq i<2^z-1$. The values of \initialBudget\ are chosen to be multiples of $\sum_{i=1}^\initialRuns l(i)$ for the \luby\ experiments and multiples of \initialRuns\ for those using \even.

Our decision makers have access to the measured data points collected until \initialBudget\ is exhausted in the form of tuples of $(\textit{time}, \textit{quality})$. The set of basic decision makers includes \currentBest, which picks the runs with the best \textit{quality} value in their last measured data point, \random, which randomly picks runs, and \currentWorst, which picks the worst runs. The latter two performed worse and were included as sanity tests only. \mostImprovements\ simply choses the run(s) $i$ with the highest number of improvements (measure points) divided by the logarithm of their consumed time \initialRunBudget{i} in the hope that they may be likely to attain further improvements. \logTimeSum\ chooses the runs for which the sum of the logarithms of all time stamps at which improvements were made are the highest.

We also propose model-based decision makers that try to construct, for each run, a functional relationship between the time stamps and the achieved quality. These relationships are used to predict the quality that a run would reach if it was selected and pick the runs with the best predicted results.

As \emph{model types} we test linear, quadratic, and cubic polynomials as well as perceptrons. The latter is suggested by \citeAsName{QWL2017MOARBAIA,QWL2018OABMASOTTSP} for modeling optimization algorithm behavior. We apply perceptrons \perceptron{n} with $n\in\{1,2,3\}$ nodes on a single hidden layer and such just with input/output layer ($n=0$). We use either $\tanh$ or the linear step as activation function. The parameters of the polynomials can either be computed directly based on two, three, or four data points or fitted using the Levenberg-Marquardt algorithm~\citep{L1944AMFTSOCNLPILS,M1963AAFLSEONP} algorithm based on last ten measured points. The parameters of the perceptrons are obtained by either applying SepCMA-ES~\citep{RH2008ASMICEALTASC} or CSA~\citep{AB2008ESWCSLAOTNPR} for at most 400 function evaluations, on the last 10 points collected in the run. We chose 10 points only in order to limit the runtime \decisionMakerTime\ consumed for training the perceptrons, which grows linearly with the number of points.

For the modeling, time and quality may be either used directly or logarithmically scaled. Furthermore, if the time value of the last measured tuple $(\textit{time}, \textit{quality})$ is less than \initialRunBudget{i}, we may add a ``virtual end point'' $(\textit{time}, \initialRunBudget{i})$ to the dataset of run $i$. This makes sense because an optimization process may first quickly trace down a local optimum and then not improve anymore at all. In that case, no further measure point would appear in its initial budget and simply extrapolating its initial progress while ignoring this fact may yield wrong predictions. Finally, we test a linear model extrapolating from the very first measured point and the ``virtual end point'' of a run into the future.

This results in 44 decision maker setups, yielding a total of $(113+86)*20*25*6*3*44=78'804'000$ experiments simulated on the data from \cite{KSW2017LARRSTISS}.%
\subsubsection{Targeted Smaller Experiment.}%
In a second experiment we investigate fewer, selected values of \initialBudget, which also allows us to test additional configurations.

We investigate one additional decision maker, \diminishingReturns, which is based on the idea of \emph{diminishing returns}~\cite{SN2001M}. We set $\Delta_q=\min\{0.95, \Delta_{q,1}/\Delta_{q,2}\}$, where $\Delta_{q,1}$ be the last improvement in terms of quality a run has made and $\Delta_{q,2}$ the previous one. We further set $\Delta_t=\max\{1.05, \Delta_{t,1}/\Delta_{t,2}\}$ where $\Delta_{t,1}$ and $\Delta_{t,2}$ are the corresponding required runtime. The decision maker assumes that it will take longer by factor $\Delta_t$ to achieve each further improvement for the run, which, in turn, will be smaller by factor $\Delta_q$. Improvements and times are always discretized.

In this experiment, we set $\selectedRuns=1$. We choose $\totalBudget\in\{2s, 10s, 20s, 50s, 100s, 200s, 500s, 1000s, 2000s, 5000s\}$ in correspondence to \cite{KSW2017LARRSTISS}, who used the range 50s to 500s for \MVC\ and 100s to 5000s for \TSP. We take 1000 samples for each setup.%
\subsection{Results}%
\subsubsection{Initial Large-Scale Experiment.}%
\begin{figure*}[t]%
\centering%
\resizebox{0.99\linewidth}{!}{%
\begin{tabular}{p{0.33\linewidth}@{\hspace{2em}}p{0.33\linewidth}@{\hspace{2em}}p{0.33\linewidth}}%
\includegraphics[width=\linewidth]{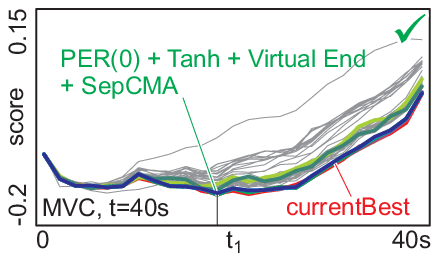}%
&%
\includegraphics[width=\linewidth]{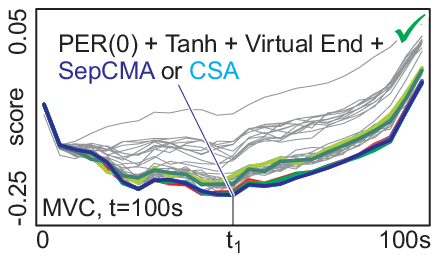}%
&%
\includegraphics[width=\linewidth]{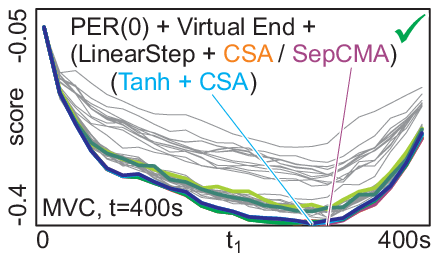}%
\\%
\multicolumn{1}{c}{\textbf{a)}~\MVC, $\totalBudget=40s$}%
&
\multicolumn{1}{c}{\textbf{b)}~\MVC, $\totalBudget=100s$}%
&%
\multicolumn{1}{c}{\textbf{c)}~\MVC, $\totalBudget=400s$}%
\medskip\\%
\includegraphics[width=\linewidth]{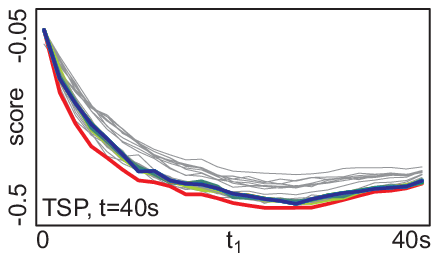}%
&%
\includegraphics[width=\linewidth]{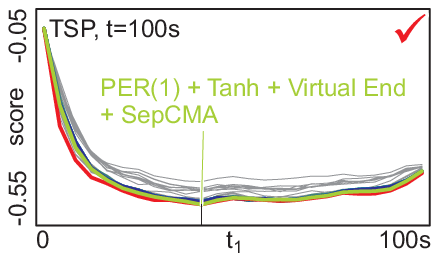}%
&%
\includegraphics[width=\linewidth]{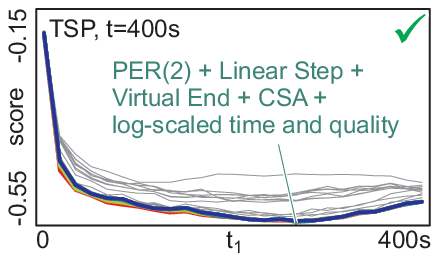}%
\\%
\multicolumn{1}{c}{\textbf{d)}~\TSP, $\totalBudget=40s$}%
&
\multicolumn{1}{c}{\textbf{e)}~\TSP, $\totalBudget=100s$}%
&%
\multicolumn{1}{c}{\textbf{f)}~\TSP, $\totalBudget=400s$}%
\\%
\end{tabular}}
\caption{Performance of the decision makers compared with a single run executed over the whole budget \totalBudget, for strategy \even, with $\initialRuns=40$, $\selectedRuns=1$, and different values of \totalBudget\ (diagrams) and \initialBudget\ (x-axes). We display the average score over all benchmark instances of the \MVC/\TSP\ datasets at the y-axes. For each time a setup yields a better result than the single run would have yielded, it receives a score of $-1$, for each time it returns a worse solution, it yields $1$ ($0$ for the same solution). Only the relevant of the 44 decision makers are highlighted. If the best-performing setup was not \currentBest, the diagram is marked with \protect\includegraphics[width=0.9em]{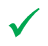}, if another setup scored equally good with \currentBest, we mark the diagram with \protect\includegraphics[width=0.9em]{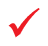}.}%
\label{fig:winLoss}%
\end{figure*}%
An experiment of the scale and with as many parameters as described in Section~\ref{sec:initialExperiment} cannot be discussed in full here. Our findings confirm that \currentBest\ is a very robust basic strategy that performs the best in many situations. We know from Section~\ref{sec:sub:selection} that good decision makers should perform very similar to it and only sometimes can yield better results. Averaged over all benchmark instances, it should be possible to gain an advantage of a few percent. From Table~\ref{tbl:scenarioChanges} we can predict that this advantage should be bigger on the \MVC\ than on the \TSP.

Indeed, in Figure~\ref{fig:winLoss}, we can observe exactly this.\footnote{%
Over all values of \initialRuns, \selectedRuns, and strategies \even\ and \luby, we can observe both scenarios where \currentBest\ is outperformed and such where it is not. We attempted to select figures without bias.%
} We find that perceptron-based decision makers work generally well and are (slightly) more likely to most-often outperform a single run with the full budget than \currentBest\ on \MVC\ for all $\initialBudget\in\left\{40s, 100s, 400s\right\}$ while this only holds for $\initialBudget=400s$ on the \TSP.

Larger total budgets \totalBudget\ seem to be beneficial when the goal is to outperform single runs or \currentBest. Note that this is a parameter which cannot be controlled by the user as it results from application requirements. 

The time \decisionMakerTime\ needed by the decision makers is generally the highest for perceptron-based methods (influenced by the presence and size of the hidden layer) and in the 100ms range. If we do not consider \decisionMakerTime\ in our simulated experiments, i.e., artificially set $\decisionMakerTime=0$, the outcome of the experiments stays almost the same. \decisionMakerTime\ is deducted from \selectedAdditionalBudget\ to be used for continuing the selected runs. It would be conceivable that using much time to make a decision could decrease \selectedAdditionalBudget\ too much so that the gain from better prediction is destroyed by the loss of budget for actually attaining the gain. However, the experiment indicates that using more complex decision makers requiring more time \decisionMakerTime\ may be viable, e.g., using more than the last 10 points to train our perceptrons would have been possible.

\begin{figure}[tb]%
\centering%
\includegraphics[width=0.9\linewidth]{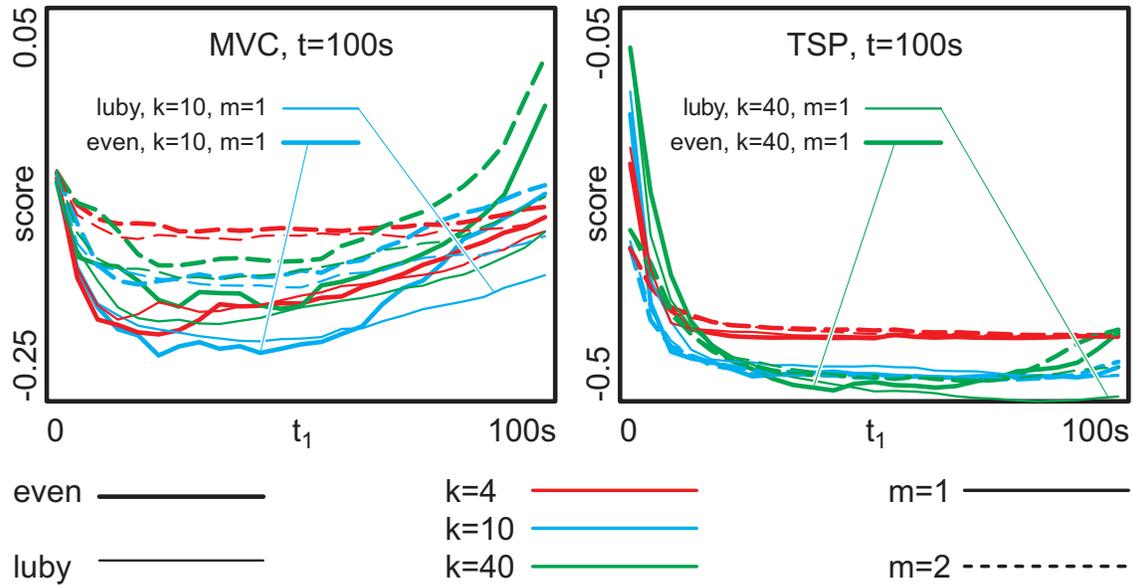}%
\caption{The average scores of the different budgeting strategies for $\totalBudget=100s$ over all decision makers and \MVC\ and \TSP\ instances. (see Figure~\ref{fig:winLoss} for definition of ``score'')}%
\label{fig:budgetingStrategies}%
\end{figure}%
We now analyze the impact of the budgeting strategy, i.e., the choices of \initialRuns, \selectedRuns, and whether to apply the \even\ or \luby\ time distributions. Good choices of these parameters should obviously depend on the available total budget \totalBudget\ and the runtime behavior of the solvers. In Figure~\ref{fig:budgetingStrategies} we plot the performance of the different configurations for $\totalBudget=100s$, averaged over \emph{all decision makers} and for different values of \initialBudget.

Using \even\ with $\initialRuns=10$, $\selectedRuns=1$ is the best choice for the \MVC\ $\totalBudget\in\{40s, 100s\}$ and the \TSP\ for $\totalBudget\in\{10s,40s\}$. For smaller budgets of the \MVC\ and $\totalBudget=100s$ on the \TSP, it makes sense to just perform $\initialRuns=40$ independent restarts distributing the time according to the \luby\ strategy. This may result from the fact that the set of decision makers over which we average also contains worse performing methods such as \currentWorst\ and \random. Continuing two runs ($\selectedRuns=2$) only is a good choice for the large budgets $t=400s$ on both the \MVC\ and \TSP. This cost of continuing a second run is only then outweighed by the benefits of exploiting the variance of runtime performance.%
%

\subsubsection{Alternatives to \currentBest.}%
As shown in Section~\ref{sec:sub:selection} and Figure~\ref{fig:winLoss}, it is theoretically and practically possible to outperform the predictor \currentBest. Next, we show to which extent and under which conditions we are able to do so given the predictors described above.

We compare our results with the best approach from \cite{FKW2017AGBARSFSUSLS} (named F17 here and used as a benchmark by us), which uses \currentBest\ to pick $\selectedRuns=1$ run from $\initialRuns=40$ initial runs, each of which received 1\% of the total time budget \totalBudget, i.e., $\initialBudget=0.4\totalBudget$. The purpose of this comparison is to see whether improvements are possible, and also whether they are statistically significant.

In Figure~\ref{fig:pies}, we show a qualitatively representative subset of our results. We have chosen two extreme total time budgets (a very small one of 2s and a very large one of 2'000s) and selected a diverse set of predictors. Note that we have chosen pie charts on purpose as they allow for a quick qualitative comparison of results.

It turns out, that the benchmark approach dominates or is dominated, depending on the problem domain, the instances, and the total time budget. For example, it is no surprise that the benchmark approach can beat the single run (lots of green) when the total time budget is large, as performance variance can be exploited. 
Also, we can see that the last phase of \bar, i.e. when a run is continued, is generally helpful when the total runtime is short, as both \even and \luby are beaten significantly and often in both the \MVC and \TSP case (lots of green). 

For \MVC and small budgets, many predictors can beat the benchmark approach. This advantage vanishes as the total time budget increases, which is due to the algorithm's convergence within the used time $\initialBudget/\initialRuns$ for the individual initial runs. Consequently, performances are typically not distinguishable anymore from F17 (lots of gray), while the differences remain statistically significant for the \TSP.

When it comes to the different problem domains, it also turns out that for \MVC many predictors perform better than the benchmark approach. For example, the  \diminishingReturns\ approach is significantly better on 43 instances while worse only on 24 instances; similar ratios hold for the other predictors. 
For the \TSP and the long total time budget, however, there are a few deviations from the ``usual'' pie chart in this category. Noteworthy deviations are the perceptrons \perceptron{0} without hidden layer and \diminishingReturns. In both cases, the benchmark is better on only three instances (as visible by the little green section), while being beaten on 19 instances (shown in red).

Lastly, we briefly compare the performance of different predictors when only 4 instead of 40 initial runs are performed. The results in Figure~\ref{fig:piesk4} show that predictors more elaborate than \currentBest\ are again significantly more successful in picking the best run for both small and large total time budgets (lots of red and gray).

\newcommand{\fancypie}[4]{%
\pgfmathsetmacro{\calcResult}{int(#1-#2-#3-#4)} 
\def\angle{0}%
\def\radius{0.7}
\def\cyclelist{{"red","green","black!15!white","black!40!white"}}%
\newcount\cyclecount \cyclecount=-1%
\newcount\ind \ind=-1%
\begin{tikzpicture}[nodes = {font=\sffamily}]%
  \foreach \percent/\name in {
      #2/worse,
      #3/better,
      #4/identical,
\calcResult/insignificant
    } {
      \ifx\percent\empty\else               
        \global\advance\cyclecount by 1     
        \global\advance\ind by 1            
        \ifnum3<\cyclecount                 
          \global\cyclecount=0              
          \global\ind=0                     
        \fi
        \pgfmathparse{\cyclelist[\the\ind]} 
        \edef\color{\pgfmathresult}         
        \draw[fill={\color!50},draw={\color}] (0,0) -- (\angle:\radius)
          arc (\angle:\angle+\percent*3.6/#1*100:\radius) -- cycle;
\ifnum5<\percent
\node at (\angle+0.5*\percent*3.6/#1*100:0.7*\radius) {\normalfont{\footnotesize{\percent}}};
\fi
        \pgfmathparse{\angle+\percent*3.6/#1*100}  
        \xdef\angle{\pgfmathresult}         
      \fi%
}%
\end{tikzpicture}%
\hspace{3mm}}


\newcommand{\labelLeft}[2]{\noindent\begin{minipage}{1.0cm}\text{#1:}\vspace*{7mm}\end{minipage}}

\begin{figure}[tp]
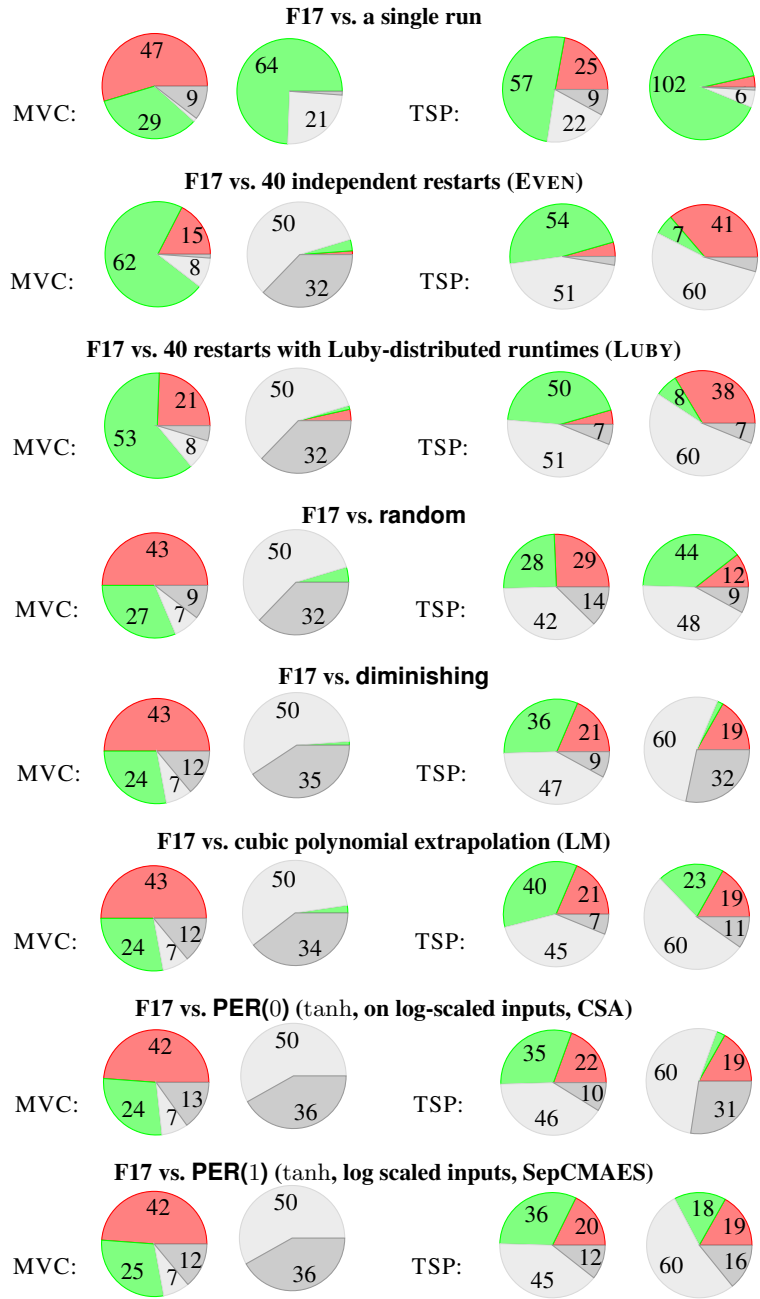

\centering\footnotesize
\textbf{F17 vs.\ a single run}\\\vspace{0mm}\labelLeft{\MVC}{gap}
\fancypie{86}{47}{29}{1}\fancypie{86}{0}{64}{21}
\hspace{5mm}\labelLeft{\TSP}{gap}
\fancypie{113}{25}{57}{22}\fancypie{113}{4}{102}{6}
\\\vspace{-1.5mm}
\textbf{F17 vs.\ 40 independent restarts (\even)}\\\vspace{0mm}\labelLeft{\MVC}{gap}
\fancypie{86}{15}{62}{8}\fancypie{86}{1}{3}{50}
\hspace{5mm}\labelLeft{\TSP}{gap}
\fancypie{113}{5}{54}{51}\fancypie{113}{41}{7}{60}
\\\vspace{-1.5mm}
\textbf{F17 vs.\ 40 restarts with Luby-distributed runtimes (\luby)}\\\vspace{0mm}\labelLeft{\MVC}{gap}
\fancypie{86}{21}{53}{8}\fancypie{86}{3}{1}{50}
\hspace{5mm}\labelLeft{\TSP}{gap}
\fancypie{113}{5}{50}{51}\fancypie{113}{38}{8}{60}
\\\vspace{-1.5mm}
\textbf{F17 vs.\ \random}\\\vspace{0mm}\labelLeft{\MVC}{gap}
\fancypie{86}{43}{27}{7}\fancypie{86}{0}{4}{50}
\hspace{5mm}\labelLeft{\TSP}{gap}
\fancypie{113}{29}{28}{42}\fancypie{113}{12}{44}{48}
\\\vspace{-1.5mm}
\textbf{F17 vs.\ \diminishingReturns}\\\vspace{0mm}\labelLeft{\MVC}{gap}
\fancypie{86}{43}{24}{7}\fancypie{86}{0}{1}{50}
\hspace{5mm}\labelLeft{\TSP}{gap}
\fancypie{113}{21}{36}{47}\fancypie{113}{19}{2}{60}
\\\vspace{-1.5mm}
\textbf{F17 vs.\ cubic polynomial extrapolation (LM)}\\\vspace{0mm}\labelLeft{\MVC}{gap}
\fancypie{86}{43}{24}{7}\fancypie{86}{0}{2}{50}
\hspace{5mm}\labelLeft{\TSP}{gap}
\fancypie{113}{21}{40}{45}\fancypie{113}{19}{23}{60}
\\\vspace{-1.5mm}
\textbf{F17 vs.\ \perceptron{0} ($\tanh$, on log-scaled inputs, CSA)%
}\\\vspace{0mm}\labelLeft{\MVC}{gap}
\fancypie{86}{42}{24}{7}\fancypie{86}{0}{0}{50}
\hspace{5mm}\labelLeft{\TSP}{gap}
\fancypie{113}{22}{35}{46}\fancypie{113}{19}{3}{60}
\\\vspace{-1.5mm}
\textbf{F17 vs.\ \perceptron{1} ($\tanh$, log scaled inputs, SepCMAES)}\\\vspace{0mm}\labelLeft{\MVC}{gap}
\fancypie{86}{42}{25}{7}\fancypie{86}{0}{0}{50}
\hspace{5mm}\labelLeft{\TSP}{gap}
\fancypie{113}{20}{36}{45}\fancypie{113}{19}{18}{60}
\\\vspace{-1.5mm}
%
%
\ignore{
\textbf{1of40, ($t_1=40\%$), recentImprovements}\\\vspace{0mm}\labelLeft{\MVC}{gap}
\fancypie{86}{43}{24}{7}\fancypie{86}{0}{3}{50}
\hspace{5mm}\labelLeft{\TSP}{gap}
\fancypie{86}{15}{35}{32}\fancypie{86}{19}{23}{33}
\\\vspace{-1.5mm}
\textbf{1of40, ($t_1=40\%$), leastOften}\\\vspace{0mm}\labelLeft{\MVC}{gap}
\fancypie{86}{44}{24}{7}\fancypie{86}{0}{0}{50}
\hspace{5mm}\labelLeft{\TSP}{gap}
\fancypie{86}{15}{31}{33}\fancypie{86}{19}{17}{33}
\\\vspace{-1.5mm}
\textbf{1of40, ($t_1=40\%$), mostImprovements}\\\vspace{0mm}\labelLeft{\MVC}{gap}
\fancypie{86}{42}{24}{7}\fancypie{86}{0}{3}{50}
\hspace{5mm}\labelLeft{\TSP}{gap}
\fancypie{86}{15}{34}{32}\fancypie{86}{19}{22}{33}
\\\vspace{-1.5mm}
\textbf{1of40, ($t_1=40\%$), currentWorst}\\\vspace{0mm}\labelLeft{\MVC}{gap}
\fancypie{86}{41}{28}{7}\fancypie{86}{0}{4}{50}
\hspace{5mm}\labelLeft{\TSP}{gap}
\fancypie{86}{13}{35}{32}\fancypie{86}{18}{26}{33}
\\\vspace{-1.5mm}
\textbf{1of40, ($t_1=40\%$), bigImprovements}\\\vspace{0mm}\labelLeft{\MVC}{gap}
\fancypie{86}{41}{28}{7}\fancypie{86}{0}{4}{50}
\hspace{5mm}\labelLeft{\TSP}{gap}
\fancypie{86}{13}{35}{32}\fancypie{86}{18}{26}{33}
\\\vspace{-1.5mm}
\textbf{1of40, ($t_1=40\%$), mlp2TanhVElogTlogQl15CMA1000}\\\vspace{0mm}\labelLeft{\MVC}{gap}
\fancypie{86}{43}{25}{7}\fancypie{86}{0}{0}{50}
\hspace{5mm}\labelLeft{\TSP}{gap}
\fancypie{86}{14}{31}{31}\fancypie{86}{19}{20}{33}
\\\vspace{-1.5mm}
\textbf{1of40, ($t_1=40\%$), modelLinearlyVElogTlogQl10}\\\vspace{0mm}\labelLeft{\MVC}{gap}
\fancypie{86}{43}{24}{7}\fancypie{86}{0}{4}{50}
\hspace{5mm}\labelLeft{\TSP}{gap}
\fancypie{86}{15}{32}{32}\fancypie{86}{18}{23}{33}
\\\vspace{-1.5mm}
\textbf{1of40, ($t_1=40\%$), modelQuadraticalyVElogTlogQl10}\\\vspace{0mm}\labelLeft{\MVC}{gap}
\fancypie{86}{44}{24}{7}\fancypie{86}{0}{4}{50}
\hspace{5mm}\labelLeft{\TSP}{gap}
\fancypie{86}{15}{33}{31}\fancypie{86}{18}{24}{33}
\\
}
%
%
%
\caption{Statistical comparison of the best \bar configuration from \protect\cite{FKW2017AGBARSFSUSLS} (here named F17) with a subset of our approaches using the Wilcoxon rank-sum test (significance level $p=0.05$) on 1'000 independent samples per setup. The approaches are compared based on the final quality gap to the best possible solution.  
Each pair of pie charts shows the outcomes for two extreme total time budgets: $\totalBudget=2s$ (left) and $\totalBudget=2'000s$ (right). In short, the more red we see, the better the alternative is compared to F17.
\newline In detail, the colors have the following meaning:
\textcolor{green}{Green} indicates that F17 is statistically better,
\textcolor{red}{Red} indicates that F17 is statistically worse,
\textcolor{black!15!white}{Light gray} indicates that both performed identically,
\textcolor{black!40!white}{Dark gray} indicates that the differences were statistically insignificant.
\newline%
All decision makers are applied for $\initialRuns=40$, $\selectedRuns=1$, $\initialBudget=0.4\totalBudget$, except for single run, \even and \luby, which have  $\initialBudget=\totalBudget$.}
\label{fig:pies}
\end{figure}

\newcommand{\fvariant}{F17$^{k=4}$\xspace}

\begin{figure}[tb]
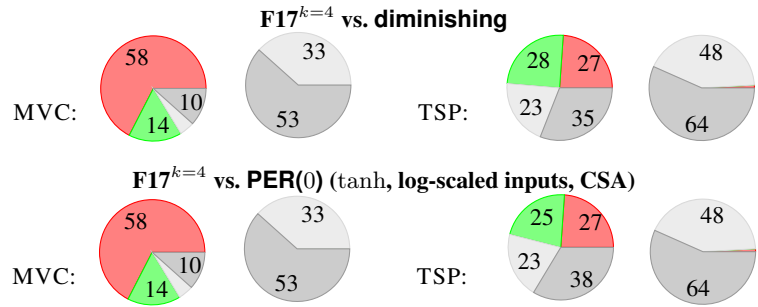

\centering\footnotesize
\textbf{\fvariant vs. \diminishingReturns}\\\vspace{0mm}
\labelLeft{\MVC}{gap}
\fancypie{86}{58}{14}{4}\fancypie{86}{0}{0}{33}
\hspace{5mm}\labelLeft{\TSP}{gap}
\fancypie{113}{27}{28}{23}\fancypie{113}{1}{0}{48}
\\\vspace{-1.5mm}
\textbf{\fvariant vs. \perceptron{0} ($\tanh$, log-scaled inputs, CSA)}\\\vspace{0mm}
\labelLeft{\MVC}{gap}
\fancypie{86}{58}{14}{4}\fancypie{86}{0}{0}{33}
\hspace{5mm}\labelLeft{\TSP}{gap}
\fancypie{113}{27}{25}{23}\fancypie{113}{1}{0}{48}
\\\vspace{-1.5mm}
%
%
%
\caption{Comparison of predictors for $\initialRuns=4$, $\selectedRuns=1$, $\initialBudget=0.4\totalBudget$. The style is identical to that of Figure~\ref{fig:pies}. As a new reference, \fvariant corresponds to the previous F17 with $k=4$.}
\label{fig:piesk4}
\end{figure}

\ignore{
\textbf{\fvariant vs. a single run}\\\vspace{0mm}
\labelLeft{\MVC}{gap}
\fancypie{86}{48}{31}{1}\fancypie{86}{0}{64}{21}
\hspace{5mm}\labelLeft{\TSP}{gap}
\fancypie{86}{16}{50}{10}\fancypie{86}{1}{70}{4}
\\\vspace{-2mm}
}

%
\section{Summary and Conclusions}%
Despite the appeal of a one-size-fits-all recommendation, we have observed in our studies that the best \bar configuration varies depending on total runtime budget \totalBudget\ and instance. The resulting ``best'' configurations have varied from ``make many short runs'' to ``just make a single run''.

This is further complicated by the observation that the very same algorithm on the very same instance can show significantly different behavior with intersecting performance profiles (see Figure~\ref{fig:10kruns}). These then cause difficulties in choosing the right run to continue: depending on the total time budget, this causes a switch of the best decision maker from ``\currentBest'' even to ``\currentWorst'' in some cases. Theoretical results are needed to characterize this further in the context of stochastic algorithms and this will be the subject of our future work.

Still, over a wide variety of scenarios, we found that predicting the future performance of the initial runs in order to select those to continue is feasible. This means that it is possible to discriminate between ``good'' and ``bad'' sample runs, and to increasing the correlation of the chosen run with the a-posteriori best one. In particular, the crude and very fast concept of diminishing returns has led to surprisingly good results. Another good approach was to fit the parameters of a perceptron to the observed data, using numerical black-box optimizers.

As our generic \bar approach is configurable, per-instance configuration should be possible. Our preliminary work in this direction indicates that this is feasible, however, the uneven heterogeneity of the instance features in combination with the small number of instances is currently posing a major challenge.

Also, the \bar approaches to date make their decisions purely based on solution quality and also consider just a single algorithm. Both aspects can be extended by giving the decision maker access to features of the solutions, and also by allowing for a diverse set of solvers (or configurations thereof) to participate in the overall optimization, with the overall goal to better exploit performance variance of solvers.


%
%
%

\theendnotes

\ACKNOWLEDGMENT{T.~Weise acknowledges support by %
the National Natural Science Foundation of China under Grants %
61673359, 
61150110488, 
and 71520107002. 
We used Alexandre Devert's great implementation of SLPs, MLPs, SepCMA, and CSA (see \url{http://github.com/marmakoide/jpack}). M.~Wagner acknowledges support by the ARC Discovery Early Career Researcher Award DE160100850.
}

\bibliographystyle{informs2014} 
\bibliography{bibliography}%

\begin{thebibliography}{40}
\providecommand{\natexlab}[1]{#1}
\providecommand{\url}[1]{\texttt{#1}}
\providecommand{\urlprefix}{URL }

\bibitem[{Abu-Khzam et~al.(2006)Abu-Khzam, Langston, Shanbhag,
  \protect\BIBand{} Symons}]{KLSS2006SPAFFP}
Abu-Khzam F, Langston M, Shanbhag P, Symons C (2006) Scalable parallel
  algorithms for {FPT} problems. \emph{Algorithmica} 45:269--284,
  \urlprefix\url{http://dx.doi.org/10.1007/s00453-006-1214-1}.

\bibitem[{Applegate et~al.(2007)Applegate, Bixby, Chv{\'{a}}tal,
  \protect\BIBand{} Cook}]{ABCC2006TTSPACS}
Applegate D, Bixby R, Chv{\'{a}}tal V, Cook W (2007) \emph{The Traveling
  Salesman Problem: A Computational Study} (Princeton University Press).

\bibitem[{Applegate et~al.(2003)Applegate, Cook, \protect\BIBand{}
  Rohe}]{ACR2003CLKFLTSP}
Applegate D, Cook W, Rohe A (2003) {Chained Lin-Kernighan} for large traveling
  salesman problems. \emph{INFORMS Journal on Computing} 15,
  \urlprefix\url{http://dx.doi.org/10.1287/ijoc.15.1.82.15157}.

\bibitem[{Arnold \protect\BIBand{} Beyer(2008)}]{AB2008ESWCSLAOTNPR}
Arnold D, Beyer H (2008) Evolution strategies with cumulative step length
  adaptation on the noisy parabolic ridge. \emph{Natural Computing} 7:555--587,
  \urlprefix\url{http://dx.doi.org/10.1007/s11047-006-9025-5}.

\bibitem[{Biere \protect\BIBand{}
  Fr\"{o}hlich(2015)}]{Biere2015evaluatingRestarts}
Biere A, Fr\"{o}hlich A (2015) Evaluating {CDCL} restart schemes.
  \emph{International Workshop on Pragmatics of SAT (POS)}, 16,
  \urlprefix\url{http://fmv.jku.at/papers/BiereFroehlich-POS15.pdf},
  preliminary version.

\bibitem[{Cai(2015)}]{C2015BBCAQLSFMVCIMG}
Cai S (2015) Balance between complexity and quality: Local search for minimum
  vertex cover in massive graphs. Yang Q, Wooldridge M, eds., \emph{Proceedings
  of the Twenty-Fourth International Joint Conference on Artificial
  Intelligence (IJCAI 2015)}, 747--753 (AAAI Press), code:
  \url{http://lcs.ios.ac.cn/~caisw/MVC.html}, accessed 2017-12-28.

\bibitem[{Cai et~al.(2015)Cai, Lin, \protect\BIBand{} Su}]{CLS2015TWLSFMVC}
Cai S, Lin J, Su K (2015) Two weighting local search for minimum vertex cover.
  \emph{29\textsuperscript{th} AAAI Conference on Artificial Intelligence (AAAI
  2015), January~25--30, 2015, Austin, United States}, 1107--1113 (AAAI Press).

\bibitem[{Cai et~al.(2013)Cai, Su, Luo, \protect\BIBand{}
  Sattar}]{CSLS2013NAELSAFMVC}
Cai S, Su K, Luo C, Sattar A (2013) Numvc: An efficient local search algorithm
  for minimum vertex cover. \emph{Journal of Artificial Intelligence Research}
  46:687--716, \urlprefix\url{http://dx.doi.org/10.1613/jair.3907}.

\bibitem[{Cir{\'{e}} et~al.(2014)Cir{\'{e}}, Kadioglu, \protect\BIBand{}
  Sellmann}]{CireKS14}
Cir{\'{e}} A, Kadioglu S, Sellmann M (2014) Parallel restarted search.
  \emph{Proceedings of the Twenty-Eighth {AAAI} Conference on Artificial
  Intelligence}, 842--848.

\bibitem[{Cook(2005)}]{C2005TTSPDW}
Cook W (2005) The traveling salesperson problem: Downloads (website).
  \url{http://www.math.uwaterloo.ca/tsp/concorde/downloads/downloads.htm},
  accessed 2017-12-28.

\bibitem[{{de Perthuis de Laillevault} et~al.(2015){de Perthuis de
  Laillevault}, Doerr, \protect\BIBand{}
  Doerr}]{dePerthuisdeLaillevault2015onemaxInits}
{de Perthuis de Laillevault} A, Doerr B, Doerr C (2015) Money for nothing:
  Speeding up evolutionary algorithms through better initialization.
  \emph{Proceedings of the Genetic and Evolutionary Computation Conference
  ({GECCO}'15), July~11-15, 2015, , Madrid, Spain}, 815--822,
  \urlprefix\url{http://dx.doi.org/10.1145/2739480.2754760}.

\bibitem[{Fischetti \protect\BIBand{} Monaci(2014)}]{FM2014EEIS}
Fischetti M, Monaci M (2014) Exploiting erraticism in search. \emph{Operations
  Research} 62:114--122,
  \urlprefix\url{http://dx.doi.org/10.1287/opre.2013.1231}.

\bibitem[{Friedrich et~al.(2017)Friedrich, K{\"o}tzing, \protect\BIBand{}
  Wagner}]{FKW2017AGBARSFSUSLS}
Friedrich T, K{\"o}tzing T, Wagner M (2017) A generic bet-and-run strategy for
  speeding up stochastic local search. Singh SP, Markovitch S, eds.,
  \emph{31\textsuperscript{st} AAAI Conference on Artificial Intelligence},
  801--807 (AAAI Press).

\bibitem[{Gagliolo \protect\BIBand{} Schmidhuber(2011)}]{Gagliolo2011}
Gagliolo M, Schmidhuber J (2011) Algorithm portfolio selection as a bandit
  problem with unbounded losses. \emph{Annals of Mathematics and Artificial
  Intelligence} 61(2):49--86,
  \urlprefix\url{http://dx.doi.org/10.1007/s10472-011-9228-z}.

\bibitem[{Gary \protect\BIBand{} Johnson(1979)}]{G1979CAIAGTTTONC}
Gary MR, Johnson DS (1979) \emph{Computers and Intractability: A Guide to the
  Theory of \npPrefix-Completeness} (New York, NY, USA: W. H. Freeman and
  Company), ISBN 0-7167-1045-5.

\bibitem[{Gomes et~al.(2000)Gomes, Selman, Crato, \protect\BIBand{}
  Kautz}]{GomesSCK00}
Gomes C, Selman B, Crato N, Kautz H (2000) Heavy-tailed phenomena in
  satisfiability and constraint satisfaction problems. \emph{Journal of
  Automated Reasoning} 24(1):67--100,
  \urlprefix\url{http://dx.doi.org/10.1023/A:1006314320276}.

\bibitem[{Gomes et~al.(2006)Gomes, {de Meneses}, Pardalos, \protect\BIBand{}
  Viana}]{GMPV2006EAOAAFTVCASCP}
Gomes F, {de Meneses} C, Pardalos P, Viana G (2006) Experimental analysis of
  approximation algorithms for the vertex cover and set covering problems.
  \emph{Computers \& OR} 33:3520--3534,
  \urlprefix\url{http://dx.doi.org/10.1016/j.cor.2005.03.030}.

\bibitem[{Gutin \protect\BIBand{} Punnen(2002)}]{GP2004TTSPAIV}
Gutin G, Punnen A, eds. (2002) \emph{The Traveling Salesman Problem and its
  Variations}, volume~12 of \emph{Combinatorial Optimization} (Berlin,
  Heidelberg: Kluwer).

\bibitem[{Huang(2007)}]{Huang2007}
Huang J (2007) The effect of restarts on the efficiency of clause learning.
  \emph{International Joint Conference on Artifical Intelligence (IJCAI)},
  2318--2323.

\bibitem[{Kadioglu et~al.(2017)Kadioglu, Sellmann, \protect\BIBand{}
  Wagner}]{KSW2017LARRSTISS}
Kadioglu S, Sellmann M, Wagner M (2017) Learning a reactive restart strategy to
  improve stochastic search. \emph{Learning and Intelligent Optimization --
  11th International Conference ({LION}~11), June~19-21, 2017, Nizhny Novgorod,
  Russia, Revised Selected Papers}, 109--123,
  \urlprefix\url{http://dx.doi.org/10.1007/978-3-319-69404-7_8}.

\bibitem[{Lalla-Ruiz \protect\BIBand{} Vo{\ss}(2016)}]{Lalla-Ruiz2016}
Lalla-Ruiz E, Vo{\ss} S (2016) Improving solver performance through redundancy.
  \emph{Systems Science and Systems Engineering} 25(3):303--325,
  \urlprefix\url{http://dx.doi.org/10.1007/s11518-016-5301-9}.

\bibitem[{Lawler et~al.(1985)Lawler, Lenstra, {Rinnooy Kan}, \protect\BIBand{}
  Shmoys}]{LLKS1985TTSPAGTOCO}
Lawler E, Lenstra J, {Rinnooy Kan} A, Shmoys D (1985) \emph{The Traveling
  Salesman Problem: A Guided Tour of Combinatorial Optimization} (New Year,
  USA: Wiley).

\bibitem[{Levenberg(1944)}]{L1944AMFTSOCNLPILS}
Levenberg K (1944) A method for the solution of certain non-linear problems in
  least squares. \emph{Quarterly of Applied Mathematics} 2:164--168.

\bibitem[{Li et~al.(2017)Li, Cai, \protect\BIBand{} Hou}]{LCH2017AELSAFMWVCOMG}
Li Y, Cai S, Hou W (2017) An efficient local search algorithm for minimum
  weighted vertex cover on massive graphs. \emph{Proceedings of the
  11\textsuperscript{th} International Conference on Simulated Evolution and
  Learning (SEAL 2017),}, 145--157,
  \urlprefix\url{http://dx.doi.org/10.1007/978-3-319-68759-9_13}.

\bibitem[{Lissovoi et~al.(2017)Lissovoi, Sudholt, Wagner, \protect\BIBand{}
  Zarges}]{Lissovoi017theory}
Lissovoi A, Sudholt D, Wagner M, Zarges C (2017) Theoretical results on
  bet-and-run as an initialisation strategy. Bosman PAN, ed., \emph{Proceedings
  of the Genetic and Evolutionary Computation Conference ({GECCO}'17),
  July~15-19, 2017, Berlin, Germany}, 857--864 ({ACM}),
  \urlprefix\url{http://dx.doi.org/10.1145/3071178.3071329}.

\bibitem[{Liu et~al.(2015)Liu, Weise, Wu, \protect\BIBand{}
  Chiong}]{LWWC2015HECMFTTSP}
Liu W, Weise T, Wu Y, Chiong R (2015) Hybrid ejection chain methods for the
  traveling salesman problem. Gong M, Pan L, Song T, Tang K, Zhang X, eds.,
  \emph{Proceedings of the 10th International Conference on Bio-Inspired
  Computing -- Theories and Applications (BIC-TA'15), September~25--28, 2015,
  Hefei, Anhui, China}, volume 562 of \emph{Communications in Computer and
  Information Science}, 268--282 (Berlin/Heidelberg: Springer-Verlag), ISBN
  978-3-662-49013-6,
  \urlprefix\url{http://dx.doi.org/10.1007/978-3-662-49014-3_25}.

\bibitem[{Louren{\c{c}}o et~al.(2010)Louren{\c{c}}o, Martin, \protect\BIBand{}
  St{\"u}tzle}]{LMS2010ILSFAA}
Louren{\c{c}}o H, Martin O, St{\"u}tzle T (2010) Iterated local search:
  Framework and applications. Gendreau M, Potvin JY, eds., \emph{Handbook of
  Metaheuristics}, volume 146 of \emph{International Series in Operations
  Research \& Management Science (ISOR)}, 363--397 (Springer),
  \urlprefix\url{http://dx.doi.org/10.1007/978-1-4419-1665-5_12}.

\bibitem[{Luby et~al.(1993)Luby, Sinclair, \protect\BIBand{}
  Zuckerman}]{LSZ1993OSOLVA}
Luby M, Sinclair A, Zuckerman S (1993) Optimal speedup of {Las Vegas}
  algorithms. \emph{Information Processing Letters} 47:173--180,
  \urlprefix\url{http://dx.doi.org/10.1016/0020-0190(93)90029-9}.

\bibitem[{Marquardt(1963)}]{M1963AAFLSEONP}
Marquardt D (1963) An algorithm for least-squares estimation of nonlinear
  parameters. \emph{SIAM Journal on Applied Mathematics} 11(2):431--441.

\bibitem[{Mart{\'i}(2003)}]{M2003MSM}
Mart{\'i} R (2003) Multi-start methods. Glover F, Kochenberger GA, eds.,
  \emph{Handbook of Metaheuristics}, volume~57 of \emph{International Series in
  Operations Research \& Management Science (ISOR)}, chapter~12, 355--368
  (Springer).

\bibitem[{Nagata \protect\BIBand{} Kobayashi(2013)}]{NK2013APGAUEACFTTSP}
Nagata Y, Kobayashi S (2013) A powerful genetic algorithm using edge assembly
  crossover for the traveling salesman problem. \emph{INFORMS Journal on
  Computing} 25:346--363,
  \urlprefix\url{http://dx.doi.org/10.1287/ijoc.1120.0506}.

\bibitem[{Qi et~al.(2017)Qi, Weise, \protect\BIBand{} Li}]{QWL2017MOARBAIA}
Qi Q, Weise T, Li B (2017) Modeling optimization algorithm runtime behavior and
  its applications. \emph{Proceedings of the Genetic and Evolutionary
  Computation Conference ({GECCO}'17) Companion, July~15-19, 2017, Berlin,
  Germany}, 115--116 (ACM),
  \urlprefix\url{http://dx.doi.org/10.1145/3067695.3076042}.

\bibitem[{Qi et~al.(2018)Qi, Weise, \protect\BIBand{} Li}]{QWL2018OABMASOTTSP}
Qi Q, Weise T, Li B (2018) Optimization algorithm behavior modeling: A study on
  the traveling salesman problem. \emph{Proceedings of the Tenth International
  Conference on Advanced Computational Intelligence (ICACI'18), March~29-31,
  2018, Xiamen, China}, 845--850 (IEEE), ISBN 978-1-5386-4362-4.

\bibitem[{Reinelt(1991)}]{R1991ATSPL}
Reinelt G (1991) {TSPLIB} {--} a traveling salesman problem library. \emph{ORSA
  Journal on Computing} 3(4):376--384,
  \urlprefix\url{http://dx.doi.org/10.1287/ijoc.3.4.376}, instances:
  \url{http://comopt.ifi.uni-heidelberg.de/software/TSPLIB95/tsp/}, accessed
  2017-12-28.

\bibitem[{Ros \protect\BIBand{} Hansen(2008)}]{RH2008ASMICEALTASC}
Ros R, Hansen N (2008) A simple modification in {CMA-ES} achieving linear time
  and space complexity. Rudolph G, Jansen T, Lucas SM, Poloni C, Beume N, eds.,
  \emph{10\textsuperscript{th} International Conference on Parallel Problem
  Solving from Nature (PPSN~X), September~13-17, 2008, Dortmund, Germany},
  volume 5199 of \emph{Lecture Notes in Computer Science}, 296--305 (Springer),
  \urlprefix\url{http://dx.doi.org/10.1007/978-3-540-87700-4_30}.

\bibitem[{Samuelson \protect\BIBand{} Nordhaus(2001)}]{SN2001M}
Samuelson P, Nordhaus W (2001) \emph{Microeconomics} (McGraw-Hill).

\bibitem[{Schoenauer et~al.(2012)Schoenauer, Teytaud, \protect\BIBand{}
  Teytaud}]{Sch-Tey-Tey:c:12}
Schoenauer M, Teytaud F, Teytaud O (2012) A rigorous runtime analysis for
  quasi-random restarts and decreasing stepsize. \emph{Artificial Evolution --
  10th International Conference, Evolution Artificielle ({EA}'11),
  October~24-26, 2011, Angers, France, Revised Selected Papers}, 37--48
  (Springer), \urlprefix\url{http://dx.doi.org/10.1007/978-3-642-35533-2_4}.

\bibitem[{Weise et~al.(2014)Weise, Chiong, Tang, L{\"{a}}ssig, Tsutsui, Chen,
  Michalewicz, \protect\BIBand{} Yao}]{WCTLTCMY2014BOAAOSFFTTSP}
Weise T, Chiong R, Tang K, L{\"{a}}ssig J, Tsutsui S, Chen W, Michalewicz Z,
  Yao X (2014) Benchmarking optimization algorithms: An open source framework
  for the traveling salesman problem. \emph{{IEEE} Computational Intelligence
  Magazine} 9(3):40--52,
  \urlprefix\url{http://dx.doi.org/10.1109/MCI.2014.2326101}.

\bibitem[{Weise \protect\BIBand{} Wagner(2018)}]{Weise2018barCode}
Weise T, Wagner M (2018) {Results of Bet-and-Run Strategies with Different
  Decision Makers on the Traveling Salesman Problem and the Minimum Vertex
  Cover Problem}. \urlprefix\url{http://dx.doi.org/10.5281/zenodo.1253770}.

\bibitem[{Whitley(2016)}]{W2016BNMDPCADIM}
Whitley D (2016) Blind no more: Deterministic partition crossover and
  deterministic improving moves. Friedrich T, Neumann F, Sutton AM, eds.,
  \emph{Companion Material Proceedings of the Genetic and Evolutionary
  Computation Conference (GECCO)}, 515--532 ({ACM}), ISBN 978-1-4503-4323-7,
  \urlprefix\url{http://dx.doi.org/10.1145/2908961.2926987}.

\end{thebibliography}
\end{document}